\documentclass[runningheads]{llncs}
\usepackage[T1]{fontenc}
\usepackage{graphicx,verbatim}
\usepackage{amsmath,amssymb,amsfonts}
\usepackage{algorithmic}
\usepackage{algorithm}
\usepackage[colorlinks,citecolor=blue,urlcolor=blue, linkcolor=blue,hypertexnames=True]{hyperref}
\begin{document}
\title{TRACE: Artifact-Robust Statistical Shape Modeling from Imperfect Surface Scans - A Case Study in Craniosynostosis 3D Photography}
\titlerunning{TRACE: Artifact-Robust Statistical Shape Modeling from Imperfect Scans}
%

\author{Sanjay Bhandari\inst{1,2} \and
Nawazish Khan\inst{1} \and
Alzbeta Novotna\inst{3} \and
Tiffany Jeong\inst{3} \and
Loretta Bowman\inst{3} \and
Michael Hernandez\inst{3} \and
Tobi Somorin\inst{3} \and
Viraj Govani\inst{3} \and
Jesse Goldstein\inst{3} \and
Shireen Elhabian\inst{1,2}}

\authorrunning{S. Bhandari et al.}

\institute{Scientific Computing and Imaging Institute, University of Utah, SLC, UT, USA \and
Kahlert School of Computing, University of Utah, Salt Lake City, UT, USA \and
Division of Pediatric Plastic Surgery, UPMC Children's Hospital of Pittsburgh, PA, USA \\
\email{sanjay.bhandari@utah.edu, shireen@sci.utah.edu}}
  
\maketitle              

\begin{abstract}
Craniosynostosis severity analysis increasingly relies on statistical shape models (SSMs) to quantify cranial morphology, but most existing workflows
depend on computed tomography or heavily curated three-dimensional (3D) photographs. Raw clinical 3D photographs provide a radiation-free and
repeatable alternative, yet often contain shoulders, hands, hair, clothing, scanner noise, and incomplete boundaries that corrupt correspondences. We introduce the Template-constrained Robust Artifact-aware Correspondence Estimation (TRACE) framework, an unsupervised method for constructing SSMs directly from artifact-contaminated clinical 3D
head photographs. TRACE predicts sparse anatomically corresponding head-surface control points from the raw point cloud, refines them through a coarse-to-fine Surface-Aware Deformation cascade, and uses thin-plate spline warping to deform a clean template mesh into a subject-specific head reconstruction. This template-constrained formulation keeps dense correspondences on clinically relevant head anatomy while suppressing non-head artifacts. The correspondence module is decoupled from the point-cloud encoder, enabling the same deformation pipeline to be paired with different backbones, including PointNet, DGCNN, and Point Transformer V3. Across all backbones, TRACE substantially improves surface sampling, topology preservation, and shape-model quality over prior SSM methods, providing a scalable foundation for photograph-based craniosynostosis shape analysis and a framework that may extend to other artifact-contaminated surface scans when an appropriate clean template is available.

\end{abstract}

\section{Introduction}
Craniosynostosis is a birth defect in which one or more of the fibrous sutures between an infant's skull bones fuse prematurely, restricting normal brain and skull growth and producing characteristic head-shape deformities~\cite{johnson2011craniosynostosis,kabbani2004craniosynostosis}. With a prevalence of roughly 1 in 2,000 to 2,500 live births~\cite{boulet2008population,timberlake2018genetics}, it is among the more common craniofacial conditions. Its clinical management is driven largely by severity, i.e., the pattern and magnitude of deformity guide surgical indication, timing, technique, and outcome assessment. Although visual inspection and manual measurements remain common, they are poorly reproducible across clinicians~\cite{kurniawan20243d}. This has motivated objective severity scoring from three-dimensional shape representations, including statistical shape models (SSMs), whose derived objective severity scores have been validated against expert craniofacial surgeon assessments and shown to correlate strongly with clinical severity ratings~\cite{anstadt2023quantifying,tao2026quantifying}. However, most quantitative severity-scoring pipelines currently rely on computed tomography (CT), which directly captures the skull bones and fused sutures that define the disease. CT-derived SSMs and related shape models therefore serve as the current reference for quantitative cranial morphology~\cite{anstadt2023quantifying,mendoza2014personalized,tao2026quantifying}. However, CT exposes infants to ionizing radiation~\cite{schaufelberger2022radiation,schweitzer2012avoiding}, a significant concern in a population that may already undergo CT for diagnosis. More importantly, CT is not well suited for frequent monitoring because of radiation-associated cancer risk, limiting population-scale longitudinal monitoring.

3D stereophotogrammetry is the practical alternative. It is fast, radiation-free, and easily repeated during routine clinic visits~\cite{abdel2023sagittal,abdel2023reliability,elkhill2023geometric,schaufelberger2022radiation}, making longitudinal and population-scale monitoring feasible in a way CT cannot. Although 3D photography captures the outer skin surface of head rather than the skull bone, Bruce et al.~\cite{bruce20233d} showed that severity scores derived from 3D stereophotogrammetry can closely match CT-derived scores for metopic craniosynostosis. That result supports 3D photography as a viable modality for craniosynostosis severity quantification. Yet their workflow required extensive manual cleaning, cropping, alignment, and landmark annotation before correspondence estimation, limiting scalability for multi-institutional and longitudinal studies.

Automated 3D photograph-based severity analysis requires valid anatomical correspondences directly from raw clinical 3D photographs, which often include shoulders, neck, clothing, hands, hair, and irregular boundaries.
Existing deep learning-based SSMs~\cite{adams2024point2ssm,adams2026point2ssm++,bhalodia2024deepssm,bhalodia2021leveraging,iyer2023mesh2ssm,sc3k} have been developed mainly for pre-segmented, artifact-free surfaces. As shown quantitatively and qualitatively in this paper, applying them directly to raw scans can place correspondence points on non-head artifacts, as illustrated in Fig.~\ref{fig:qualitative}, mixing anatomy with acquisition variation. The key technical gap is therefore not surface fitting alone, but learning anatomically meaningful head-surface correspondences from artifact-contaminated clinical photographs.

Thus, we introduce the Template-constrained Robust Artifact-aware Correspondence Estimation (TRACE) framework, an unsupervised method for building SSMs directly from raw clinical 3D head photographs, avoiding the need for heavy manual cleaning. Instead of learning correspondences on the cropped and cleaned scan, TRACE predicts sparse control points for the head surface on the given artifact-contaminated 3D photograph and uses them to drive a thin-plate spline (TPS) deformation of a clean head template. This constrains the dense correspondences to clinically relevant head anatomy while suppressing shoulders, clothing, hands, hair, and scanner artifacts. We instantiate the same correspondence-and-deformation framework with PointNet~\cite{pointnet}, DGCNN~\cite{dgae}, and Point Transformer V3 (PTv3)~\cite{ptv3} backbones, denoted PN-TRACE, DG-TRACE, and PT-TRACE. These encoders represent distinct point-cloud learning paradigms, yet all three TRACE variants substantially outperform the evaluated prior SSMs and perform similarly to one another. This suggests that the performance gain comes primarily from the proposed template-constrained correspondence framework rather than from a particular encoder, while leaving room for future improvements from advances in point-cloud representation learning. Establishing high-quality SSMs from raw 3D photographs is a necessary prerequisite for extending quantitative severity scoring to this modality at scale, and this paper addresses that prerequisite. In summary, our main contributions are:
\begin{itemize}
    \item We propose TRACE, an unsupervised SSM framework for craniosynostosis analysis directly from noisy
    clinical 3D head photographs, reducing the manual preprocessing that has so
    far limited photograph-based severity scoring to small, hand-curated cohorts.
    \item We introduce an artifact-robust correspondence-and-deformation
    strategy that predicts sparse head-surface control points and uses them to
    drive a TPS deformation of a clean template mesh, thereby reconstructing
    subject-specific head anatomy while suppressing non-head artifacts
    such as shoulders, hands, clothing, and scanner noise.
    \item We use TRACE's modular design to evaluate the same
    correspondence-and-deformation framework with PointNet, DGCNN, and Point
    Transformer V3 backbones, showing that the performance gain is attributable to the framework rather than to a specific point-cloud encoder.
    \item We demonstrate through our experiments that TRACE produces more accurate surface sampling, better topology preservation, and higher-quality statistical shape model than prior deep learning-based SSMs.
\end{itemize}

\section{Literature Review}
Statistical shape models (SSMs) provide a compact and interpretable representation of anatomical variation by establishing dense correspondences across a population and applying Principal Component Analysis (PCA) to the resulting shape descriptors. In craniosynostosis, SSMs have been used to quantify severity by comparing patients against normative atlases~\cite{mendoza2014personalized}, projecting morphology into demographic normative spaces~\cite{elkhill2023geometric}, and deriving severity scores from PCA modes~\cite{rodriguez2017quantifying}. These applications depend directly on the anatomical consistency of the underlying correspondences. Constructing trustworthy SSMs from raw 3D photographs is thus a prerequisite for automated photograph-based severity analysis.

Recent deep learning methods have made substantial progress in point-cloud representation learning and correspondence estimation. PointNet~\cite{pointnet} uses shared multi-layer perceptrons and max pooling to learn permutation-invariant point-cloud features, while DGCNN~\cite{dgae} builds dynamic nearest-neighbor graphs and applies edge convolution to capture local geometric structure. Transformer-based models, including Point Transformer~\cite{pt}, Point Transformer V2~\cite{ptv2}, and Point Transformer V3~\cite{ptv3}, further improve point-cloud representation learning through attention-based feature aggregation. These backbones have enabled a range of learned correspondence and shape-modeling approaches. Achlioptas et al.~\cite{pnae} showed that PointNet-based autoencoders can learn useful latent shape representations, and Adams et al.~\cite{adams2023can} showed that SSMs can be derived from such learned representations. Deep Point Correspondence (DPC)~\cite{dpc} establishes correspondence by using latent similarity to reorder a source point cloud to match a target point cloud. Chen et al.~\cite{isr} proposed intrinsic structural representation (ISR) points using a PointNet++~\cite{pointnet++} encoder and an MLP-based point integration module. Bhalodia et al.~\cite{bhalodia2021leveraging} learned anatomically corresponding landmarks through a self-supervised image registration framework, using the predicted landmarks as control points for a TPS transformation~\cite{duchon2006splines,keller2019thin} before applying PCA to construct an SSM. Mesh2SSM~\cite{iyer2023mesh2ssm} learns anatomically consistent correspondences through template deformation and then models nonlinear population variation with a variational autoencoder. SC3K~\cite{sc3k} discovers semantically consistent 3D keypoints from point clouds while remaining robust to rotation, noise, and downsampling. Point2SSM~\cite{adams2024point2ssm} predicts dense anatomically corresponding surface points from raw point clouds using a DGCNN-based attention network and applies PCA to those correspondences. Point2SSM++~\cite{adams2026point2ssm++} extends this approach with consistency learning to enforce sampling invariance and rotation equivariance.

These methods are evaluated mainly on pre-segmented, artifact-free surfaces. On raw artifact-contaminated 3D photographs, they can encode full-scan variation rather than isolating head anatomy. TRACE addresses this limitation with a template-constrained correspondence-and-deformation framework that restricts the learned shape model to the head surface.

\section{Method}
\label{sec:method}
\subsection{Problem Formulation and Overview}
\label{sec:overview}

\begin{figure}[t]
\includegraphics[trim = 0cm 1cm 0cm 1cm, width=\linewidth]{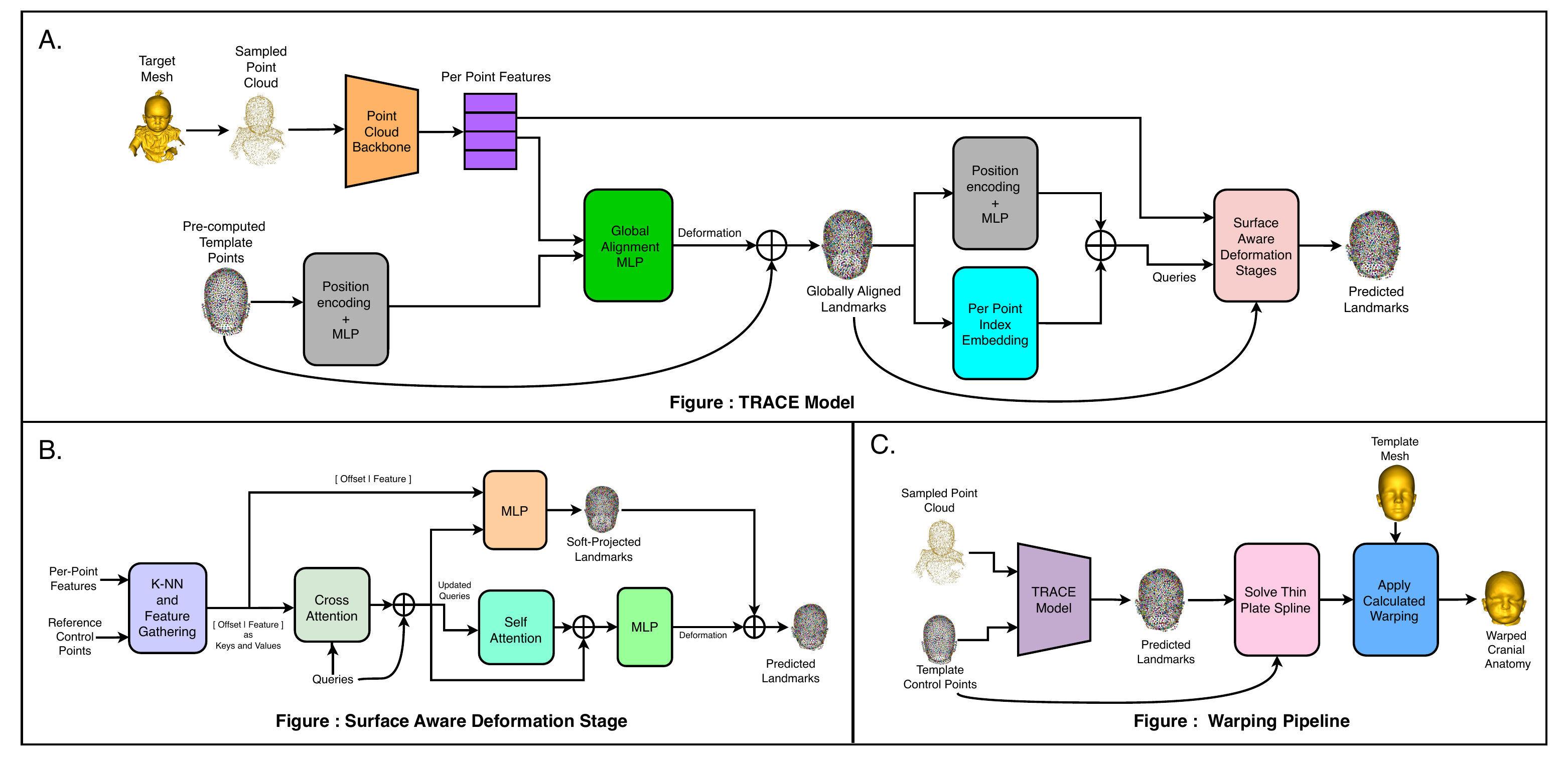}
\centering
\caption{Overview of the Template-constrained Robust Artifact-aware Correspondence Estimation (TRACE) framework. \textbf{(A)} A point-cloud backbone encodes the input scan, global-alignment stage predicts initial template-control-point displacements, and two Surface-Aware Deformation (SAD) stages refine landmarks onto the observed head surface. \textbf{(B)} A SAD stage takes the current reference landmarks and query embeddings. For each landmark, it gathers the $k_s$ nearest target points and backbone features, uses cross-attention to form a neighborhood-aware query, softly projects the landmark as a weighted average of nearby target points, and applies self-attention plus an MLP to predict a residual update. \textbf{(C)} TPS warping deforms the dense template mesh using the transformation defined from template control points toward predicted landmarks, reconstructing head anatomy while suppressing artifacts.}
\label{fig:TRACE}
\end{figure}

We consider unsupervised anatomical landmark prediction on 3D head-surface point clouds. Let $\mathcal{X}=\{X_1,X_2,\ldots,X_N\}$ denote a set of aligned clinical surface meshes, where $X_i=(V_i,E_i)$ is defined by vertices $V_i$ and edges $E_i$. Because scans have a variable number of vertices and may include non-cranial artifacts such as shoulders and hands, we uniformly sample a fixed-size input point cloud $P_i=\{p_m\}_{m=1}^{M}$, with $M<|V_i|$, while using $V_i$ as the deformation target. We use control points, landmarks, and correspondence points interchangeably for the sparse anatomical point representation. 

Given $P_i$, TRACE predicts $K$ subject-specific head-surface landmarks and uses them to deform a clean, topologically consistent head-surface template mesh. As shown in Fig.~\ref{fig:TRACE}, the pipeline includes: (i) a global-alignment stage that produces an initial landmark configuration; (ii) a cascade of $S$ Surface-Aware Deformation (SAD) stages that progressively refine landmarks onto the observed head surface; and (iii) a thin-plate spline (TPS) warping stage that propagates the landmark deformation to the dense template mesh. The correspondence-prediction module is decoupled from the backbone, so any permutation-invariant encoder can be substituted without changing the deformation pipeline. Although we evaluate in the context of craniosynostosis shape analysis, the framework is designed to be anatomy-agnostic: the template is the only anatomy-specific component, and substituting it with a clean mesh of any target anatomy directly extends the pipeline to new clinical settings.

\subsection{Template-constrained Robust Artifact-aware Correspondence Estimation}
\subsubsection{Feature initialization and Global Landmark Alignment:}
\label{sec:alignment}
Each input point cloud is encoded by a point-cloud backbone into per-point features $F_i=\{f_m\}_{m=1}^{M}$. A fixed set of pre-computed template control points $C=\{c_j\}_{j=1}^{K}$ is encoded with a positional encoding followed by an MLP. The per-point backbone features $F_i$ are aggregated by mean pooling into a global shape descriptor, which is concatenated with each encoded template control point. A global-alignment MLP then predicts an initial displacement $\Delta c_j$ for every
control point, yielding the globally aligned landmark:
\begin{equation}
a_j = c_j + \Delta c_j , \qquad j=1,\ldots,K .
\end{equation}
The resulting set $A_i=\{a_{ij}\}_{j=1}^{K}$ provides a globally aligned landmark configuration for subject i. This stage captures the coarse scale and global head shape, but it does not fully resolve fine surface-level correspondence. Therefore, $A_i$ is used as the set of reference landmarks for the first SAD stage. The corresponding per-control-point features initialize the first-stage query embeddings $Q_i=\{q_{ij}\}_{j=1}^{K}$, obtained by summing the positional encoding and the per-point index embedding.

\subsubsection{Surface-Aware Deformation Stage:}
\label{sec:deformation}
Each Surface-Aware Deformation(SAD) stage refines reference landmarks $R_i=\{r_{ij}\}_{j=1}^{K}$ by first projecting them toward the observed head surface and then applying an attention-based residual deformation. For each input point cloud $P_i$, we compute the head-region bounding box diagonal, $\delta_i$, from $R_i$, and define the stage-specific neighborhood radius as: $\eta_i = \rho_s * \delta_i$, where $\rho_s$ is the scaling factor that controls the spatial scale of each landmark for an SAD stage. For each reference landmark, we gather the $k_s$ nearest target points $\mathcal{N}_j=\{p_{jn}\}_{n=1}^{k_s}$ within $\eta_i$, along with features $\{f_{jn}\}_{n=1}^{k_s}$. To make the local geometry scale-invariant, we normalize each neighbor offset by $\eta_i$, and form a per-neighbor descriptor, $d_{jn} = \Big[\, \big(p_{jn} - r_j\big)/\eta_i,\; f_{jn} \,\Big]$. These descriptors serve as keys and values that the query $q_j$ attends to, in a cross-attention block, yielding an updated query $\hat{q}_j$. For each neighbor, an MLP maps the concatenation $[\hat{q}_j,\, d_{jn}]$ to a scalar logit, and a softmax over the $k$ neighbors produces soft projection weights $w_{jn}$ satisfying $\sum_{n=1}^{k_s} w_{jn}=1$. The landmark is softly projected onto the target surface as a convex combination of its neighbors, defined as:
\begin{equation}
    \tilde{z}_j = \sum_{n=1}^{k_s} w_{jn}\, p_{jn}
\end{equation}

The updated queries are then passed through a self-attention block to model dependencies among landmarks, and a final MLP predicts a residual deformation
$\Delta z_j$. The stage output is denoted as $Z_i=\{z_{ij}\}_{j=1}^{K}$, with each landmark passed to the next stage as:
\begin{equation}
z_j = \tilde{z}_j + \Delta z_j .
\end{equation}

Our TRACE model uses two SAD stages to separate coarse anatomical placement from local surface localization. The first uses globally aligned landmarks with a larger radius and neighborhood to recover from residual scale and shape mismatch. The second uses first-stage predictions as anatomically anchored references and searches within a smaller neighborhood to sharpen localization, correct residual offsets, and preserve template ordering. The ablation experiments evaluate this coarse-to-fine design.

\subsection{Template Warping via Thin-Plate Spline}
\label{sec:warping}

Following Bhalodia et al.~\cite{bhalodia2021leveraging}, we estimate a thin-plate spline transformation $\mathcal{T}_{\mathrm{TPS}}$ that maps the
template control points $C$ to the final predicted correspondence ${Z}$. Their framework applies the TPS in an image-to-image registration setting, where both the source and target control points are predicted directly on the clean images being registered. TRACE instead fixes the source side of the correspondence, i.e., the control points $C$ come from a pre-computed, artifact-free template, and only the target landmarks $Z$ are predicted on the artifact-contaminated point cloud. This  lets the warping reconstruct subject-specific head anatomy while remaining largely insensitive to non-head structures present in the raw scan. Applying $\mathcal{T}_{\mathrm{TPS}}$ to the dense template vertices $V_T$ yields the
subject-specific head surface reconstruction,
\begin{equation}
\hat{V}_i = \mathcal{T}_{\mathrm{TPS}}(V_T).
\end{equation}
Because the deformation is determined solely by the predicted landmarks, which the model places on the head surface, artifacts have minimal influence on the reconstruction. The warped mesh, therefore, preserves the template topology while adapting its geometry to the subject's head surface. Although TPS provides a smooth interpolation between the template control points and the predicted landmarks, very large or highly non-uniform landmark displacements can lead to substantial local stretching and distortion of the warped surface, particularly in regions where control-point support is sparse.

\subsection{Loss Functions}
\label{sec:losses}

TRACE is trained without correspondence annotations, with all losses evaluated in a per-shape normalized frame. The template control points are projected to their nearest target-surface points, and the centroid and bounding box diagonal of these projections define the normalization for each subject. We denote the normalized predicted landmarks, normalized template control points, normalized target points, and normalized warped template vertices as $\bar{z}$, $\bar{c}$, $\bar{t}$, and $\bar{v}$, respectively, and the corresponding sets as  $\bar{T}=\{\bar{t}\}$, and  $\bar{V}=\{\bar{v}\}$. The individual losses are defined below: 
 
\noindent{\textbf{Surface-sampling losses:}}
We define the point-to-surface (P2S) loss $\mathcal{L}_{\mathrm{ps}}$ and the warping loss $\mathcal{L}_{\mathrm{warp}}$ as single-directional Chamfer distances from the predicted landmarks, and the TPS-warped mesh to the target surface, respectively. We use single-directional rather than bidirectional Chamfer distances because the landmarks should lie on the target head anatomy, and the warped template should recover only the target head anatomy. Overall, the losses are:
\begin{equation}
\label{surfacesample}
\mathcal{L}_{\mathrm{ps}}=\frac{1}{K}\sum_{j=1}^{K}\min_{\bar{t}\in \bar{T}}
\|\bar{z}_j-\bar{t}\|_2^2,
\qquad
\mathcal{L}_{\mathrm{warp}}=\frac{1}{|\bar{V}|}\sum_{\bar{v}\in\bar{V}}
\min_{\bar{t}\in \bar{T}}\|\bar{v}-\bar{t}\|_2^2 .
\end{equation}

\noindent{\textbf{Topology-preserving loss:}}
To preserve the local geometric topology during deformation, we enforce edge-length consistency between the predicted landmarks $\bar{z}$ and the fixed template control points $\bar{c}$. For each template control point $\bar{c}_j$, let $\mathcal{N}_j^{\mathrm{tpl}}$ denote its $k_{tpl}$ nearest neighbors pre-computed in the fixed template space. The localized edge-length vectors for the predicted and template configurations are $ \mathbf{e}^{z}_j = \big(\|\bar{z}_{\ell}-\bar{z}_{j}\|_2\big)_{\ell\in
\mathcal{N}_j^{\mathrm{tpl}}} $ and 
$\mathbf{e}^{c}_j = \big(\|\bar{c}_{\ell}-\bar{c}_{j}\|_2\big)_{\ell\in
\mathcal{N}_j^{\mathrm{tpl}}} $.
With $\mathcal{S}_{L_1}(\cdot,\cdot)$ as the element-wise smooth-$L_1$ loss, the topology-preserving loss over all $K$ control points is:
\begin{equation}
\label{topo}
\mathcal{L}_{\mathrm{topo}} =
\frac{1}{K}\sum_{j=1}^{K}
\mathcal{S}_{L_1}\!\left(\mathbf{e}^{z}_j, \mathbf{e}^{c}_j\right)
.
\end{equation}

\noindent{\textbf{Repulsion loss:}}
To prevent the predicted landmarks from clustering at a single location and to encourage uniform spatial coverage, we apply a repulsion loss defined as:
\begin{equation}
\mathcal{L}_{\mathrm{rep}}=
\frac{1}{K(K-1)}\sum_{i=1}^{K}\sum_{\substack{j=1\\ j\ne i}}^{K}
\exp\!\left(-\frac{\|\bar{z}_i-\bar{z}_j\|_2^2}{2\sigma^2}\right).
\end{equation}

\noindent{\textbf{Sampling-consistency loss:}}
To ensure the network outputs the same landmark configuration for two independent samplings of the same target mesh, yielding normalized predictions
$\bar{z}^{(1)}$ and $\bar{z}^{(2)}$, we apply a sampling-consistency loss given by:
\begin{equation}
\mathcal{L}_{\mathrm{samp}}=\frac{1}{K}\sum_{j=1}^{K}
\|\bar{z}^{(1)}_j-\bar{z}^{(2)}_j\|_2^2 .
\end{equation}

\noindent{\textbf{Total objective: }}
For each prediction stage, these terms are combined as
\begin{equation}
\mathcal{L}_{\mathrm{stage}} =
\lambda_{\mathrm{ps}}\mathcal{L}_{\mathrm{ps}}+
\lambda_{\mathrm{warp}}\mathcal{L}_{\mathrm{warp}}+
\lambda_{\mathrm{topo}}\mathcal{L}_{\mathrm{topo}}+
\lambda_{\mathrm{rep}}\mathcal{L}_{\mathrm{rep}}+
\lambda_{\mathrm{samp}}\mathcal{L}_{\mathrm{samp}} ,
\end{equation}
where $\lambda_{\mathrm{ps}}$, $\lambda_{\mathrm{warp}}$, $\lambda_{\mathrm{topo}}$,
$\lambda_{\mathrm{rep}}$, and $\lambda_{\mathrm{samp}}$ are non-negative weights controlling the relative contribution of the point-to-surface, warping, topology, repulsion, and sampling-consistency terms, and these weights sum to $1.0$. Losses are evaluated at the
global-alignment output and at each SAD stage output and summed together: 
\begin{equation}
\mathcal{L}
=
\omega_0 \mathcal{L}_{\mathrm{global}}
+
\sum_{s=1}^{S}
\omega_s \mathcal{L}_{\mathrm{SAD}}^{(s)},
\end{equation}
where $\omega_0$ and $\omega_s$ denote the stage weights for the global-alignment output and the SAD stage outputs, respectively.

The warping loss encourages TRACE to explain each subject through deformation of the clean template mesh, rather than by directly reconstructing all structures present in the raw scan. Because the template contains only the head surface, this loss is designed to encourage the predicted control points to explain anatomically relevant head regions of the target mesh rather than non-head artifacts. The topology-preservation and repulsion losses further regularize the predicted landmarks by maintaining the local geometric organization of the template and preventing landmark collapse, thereby reducing the likelihood that correspondences drift toward isolated artifacts or dense non-head structures. In addition, the fixed template control points provide a strong anatomical prior, i.e., the global-alignment stage begins from a plausible head-shaped configuration instead of unconstrained points distributed over the full clinical scan. The coarse-to-fine SAD cascade then further enforces encoding of head structure through local surface refinement around anatomically anchored reference landmarks.

\section{Experiments}
The dataset used in this study was made available through the consortium sites as part of the CranioRate framework~\cite{craniorate}. It comprises 201 3D surface meshes reconstructed from 3D photographs of patients spanning a range of ages, treatment statuses, and craniofacial phenotypes. Since each raw mesh contains a variable number of vertices, we randomly sample 5000 vertices uniformly from each mesh to serve as the input point cloud. The data are partitioned into 151 samples for training, 25 for validation, and 25 for testing. Since some scans contain artifacts, meshes are manually aligned around the center of the head structure, rather than the full mesh, before training. 

TRACE uses a mean outer-head-surface template obtained from the CraniumPy toolbox~\cite{abdel2023reliability,abdel2023sagittal,craniumpy}, derived from normocephalic infant 3D stereophotogrammetry and based on the public SSM of Schaufelberger et al.~\cite{schaufelberger2022radiation}.
The pipeline operates entirely on head-surface data, with no CT dependency.
We pre-compute 2048 template control points using ShapeWorksStudio~\cite{cates2017shapeworks} as source points for correspondence prediction and TPS warping, thus all models also predict 2048 corresponding control points for target surface. All baselines use the same splits, comparable validation tuning, and the same number of correspondence points. TRACE estimates TPS from pre-computed template control points to predicted target landmarks, whereas baselines estimate TPS from model-predicted template points to model-predicted target points on the noisy scan.

The deformation module has two SAD stages, each defined with neighborhood and radius settings as $(k_s=64, \rho_s=0.36)$ and $(k_s=16, \rho_s=0.09)$, and topology loss uses $k_{tpl}=8$ template neighbors. All models are trained for 200 epochs on an NVIDIA H200 GPU with batch size 4 and initial learning rate $10^{-3}$. After tuning on the validation set, we set the weight of global-alignment stage loss, $\omega_0$ to 0.1, the weight of each SAD stage loss, $\omega_s$ to 1.0, and the individual loss weights are set to $\lambda_{\mathrm{ps}}=0.5$, $\lambda_{\mathrm{warp}}=0.1$, $\lambda_{\mathrm{topo}}=0.25$, $\lambda_{\mathrm{rep}}=0.1$, and $\lambda_{\mathrm{samp}}=0.05$. 

Because this paper is a method contribution focused on automated SSM construction from raw clinical 3D photographs, we evaluate the learned correspondences using established SSM quality criteria rather than downstream clinical severity scores. Compactness, generalization, and specificity are well-established evaluation metrics in the SSM literature~\cite{cates2014computational} and directly measure whether the learned correspondences form a compact, generalizable, and anatomically plausible population shape space. These metrics therefore provide the appropriate validation target for the present work, whose goal is to establish the automated shape-modeling foundation required before large-scale severity scoring can be clinically validated.

\section{Results}
We instantiate TRACE with DGCNN~\cite{dgae}, PointNet~\cite{pointnet}, and Point Transformer V3 (PTv3)~\cite{ptv3} backbones, denoted as DG-TRACE, PN-TRACE, and PT-TRACE respectively. We compare against PN-AE~\cite{pnae}, DG-AE~\cite{dgae}, ISR~\cite{isr}, DPC~\cite{dpc}, CPAE~\cite{cpae}, Point2SSM~\cite{adams2024point2ssm}, and Point2SSM++~\cite{adams2026point2ssm++}. We report surface sampling, topology preservation, and statistical shape-model quality. Surface sampling uses point-to-face distance (P2F) for predicted landmarks and surface-to-surface distance (S2S) for the TPS-warped template. 
P2F is computed as the average Euclidean distance from each predicted correspondence point to the closest face on the dense target mesh. S2S is computed as the average Euclidean distance from each point on the TPS-warped template mesh to the closest point on the dense target mesh. The Topology metric is measured with Eq.~\ref{topo}, which compares local edge-length relationships of predicted points with the pre-computed template points. We note that this topology metric coincides with a topology-preserving loss used by TRACE, whereas the other evaluated methods do not explicitly optimize this objective during training. To evaluate the quality of the learned statistical shape model, we construct an SSM from the correspondence points predicted by each method, following the approach of Cates et al.~\cite{cates2014computational}.
Shape-model quality is measured by PCA compactness, generalization, and specificity. Compactness is the number of PCA modes required to explain 95\% of the population-level shape variation. Generalization measures how accurately the shape model reconstructs held-out shapes after projecting them into the learned PCA space. Specificity measures how closely shapes sampled from the PCA model resemble realistic shapes in the dataset, calculated by sampling from the PCA distribution and measuring their distances to the closest real shape in the dataset. All distance metrics are reported in the original data space, and lower is better.

\begin{figure}[t]
\centering
\includegraphics[width=\textwidth]{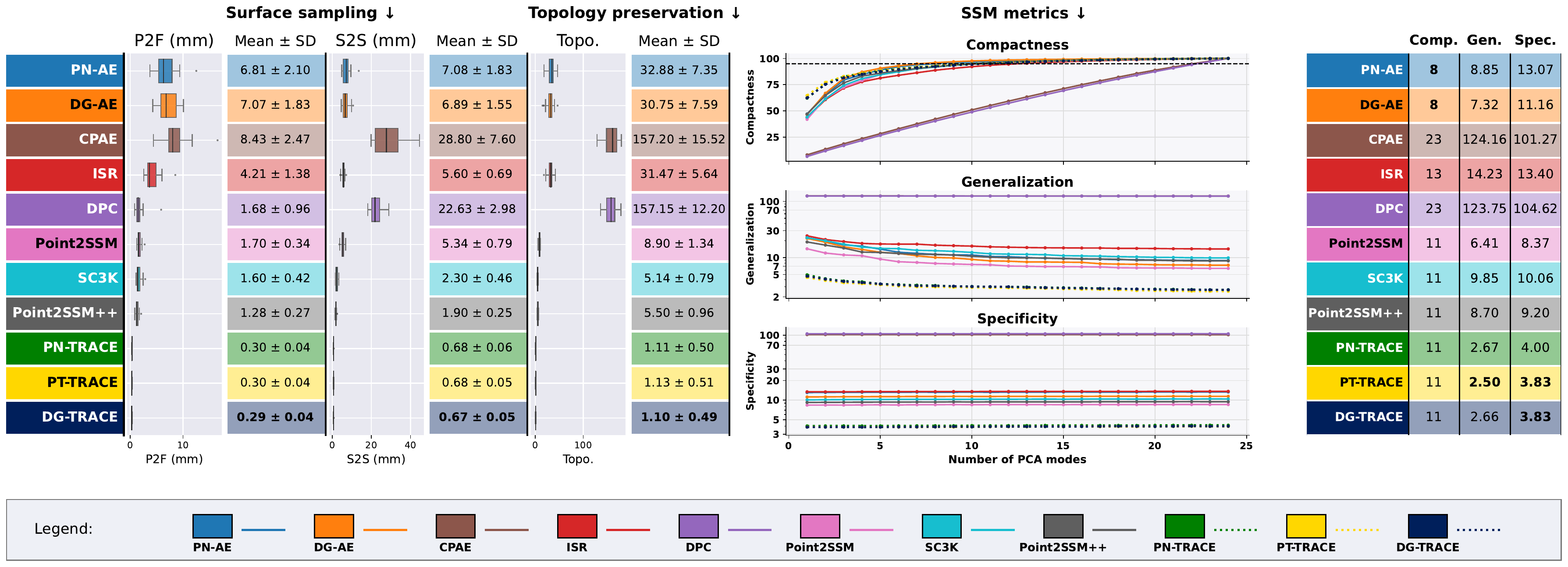}
\caption{ Quantitative comparison of our modular architecture against prior SSMs on the held-out test set. We report point-to-face distance (P2F), surface-to-surface  distance (S2S), and topology preservation metric (Topo), as well as the statistical shape model metrics, compactness (Comp.), generalization (Gen.), and specificity (Spec.). Lower is better for all metrics. Bold characters denote the best quantitative result for a metric. }
\label{fig:quantitative}
\end{figure}

\subsection{Quantitative Results}
Fig.~\ref{fig:quantitative} shows that all TRACE variants substantially improve average surface sampling and topology preservation metrics over prior methods. Point2SSM++, the strongest baseline, obtains 1.28 mm P2F and 1.90 mm S2S, whereas TRACE reduces these average distances to 0.29 to 0.30 mm and 0.67 to 0.68 mm, respectively. TRACE also reduces the average topology error to 1.10 to 1.13, compared with the best baseline value of 5.14. Among the TRACE variants, DG-TRACE achieves the lowest average surface errors, with 0.29 mm P2F and 0.67 mm S2S, and a topology error of 1.10. Methods with lower topology error generally also achieve better generalization and specificity. PN-AE and DG-AE require the fewest PCA modes to explain 95\% of variation, but their higher generalization and specificity errors indicate that this compactness reflects a restricted or less representative shape space rather than better anatomical correspondence. In contrast, all TRACE variants require 11 modes, comparable to the strongest correspondence-based baselines, while achieving the lowest generalization and specificity errors. PT-TRACE obtains the best generalization score of 2.50, and PT-TRACE and DG-TRACE achieve the best specificity score of 3.83. Together, these results indicate that the cleaner, artifact-robust correspondences produced by TRACE yield population shape spaces that better reconstruct unseen head anatomy and generate samples closer to realistic anatomical shapes. This property is particularly important for downstream craniosynostosis analysis, where severity measures should reflect true head morphology rather than acquisition artifacts.

\begin{figure}[t]
\centering
\includegraphics[width=\textwidth]{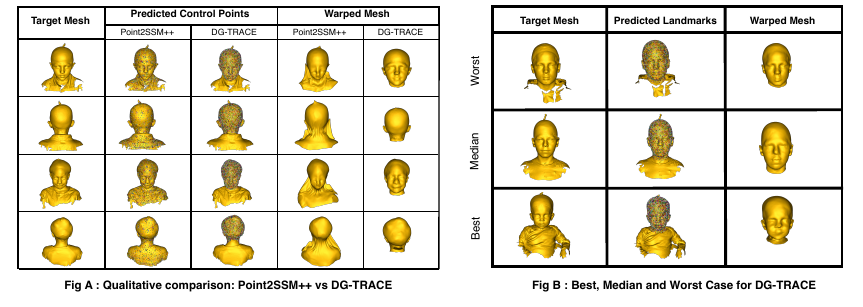}
\caption{ \textbf{Qualitative Results:} \textbf{(A)} Qualitative comparison between Point2SSM++ and DG-TRACE. \textbf{(B)} Best-performing, median-performing, and worst-performing DG-TRACE test cases ranked by the sum of P2F, S2S, and topology
error.}
\label{fig:qualitative}
\end{figure}

\begin{figure}[t]
\centering
\includegraphics[width=\textwidth]{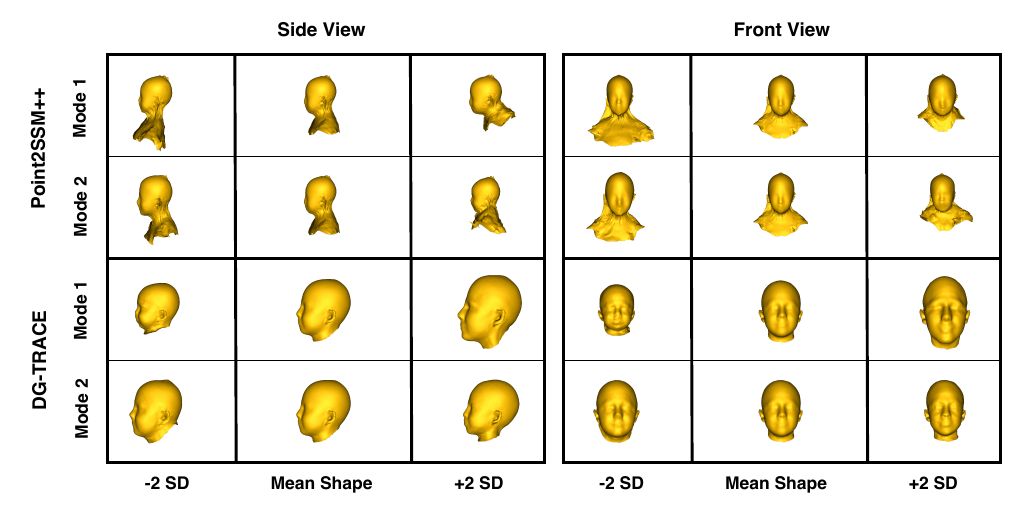}
\caption{ \textbf{Shape Model:} Side-view and Facial front view of the first two PCA modes of population shape variation for Point2SSM++ and DG-TRACE.}
\label{fig:pca}
\end{figure}

\subsection{Qualitative Results}
We qualitatively compare DG-TRACE with Point2SSM++. Point2SSM++ is the strongest surface-sampling baseline and remains competitive on SSM metrics, while DG-TRACE has the best TRACE surface-sampling performance with SSM quality comparable to PN-TRACE and PT-TRACE.

\noindent{\textbf{Surface sampling:}}
Fig.~\ref{fig:qualitative}(A) compares front and rear views of two representative scans with common 3D photography artifacts. Point2SSM++ predicts control points on the shoulders and lower neck along with head structure. Consequently, the Point2SSM++ warped meshes preserve artifact structures too, and fail to preserve fine head structures reliably. DG-TRACE instead keeps control points on the head surface, reconstructing the head while suppressing lower-surface artifacts. The resulting meshes preserve the head surface and facial contour, including in irregular cases, but still do not fully preserve fine-grained structures such as the eyes and mouth. This matches the quantitative gains in P2F, S2S, and topology preservation, indicating that the learned correspondences are not only close to the observed surface but also anatomically meaningful for craniosynostosis shape analysis. Fig.~\ref{fig:qualitative}(B) further summarizes DG-TRACE behavior across the held-out test set by showing the best-performing, median-performing, and worst-performing cases ranked by the sum of P2F, S2S, and topology error. The best case has a combined error of 1.234, i.e., P2F of 0.277 mm, S2S of 0.638 mm, Topo of 0.319, with low surface distances and strong topology preservation. The median case has a combined error of 2.194, i.e., P2F of 0.295 mm, S2S of 0.697 mm, Topo of 1.203, and the worst case has a combined error of 3.135, i.e., P2F of 0.289 mm, S2S of 0.710 mm, Topo of 2.136, where the increase is driven primarily by loss of local correspondence organization rather than by a large surface-distance failure alone. Even in the worst case, DG-TRACE produces a coherent head reconstruction, which is an important prerequisite for future scalable craniosynostosis severity evaluation. 

\noindent{\textbf{Population shape space:}}
Fig.~\ref{fig:pca} compares the first two PCA modes from Point2SSM++ and DG-TRACE correspondences. Point2SSM++ modes mix head and facial variation with neck, shoulder, and lower-surface artifacts, reducing anatomical interpretability and potentially corrupting severity analysis. DG-TRACE modes remain localized to head anatomy and capture interpretable variation, i.e., the first mode reflects global head scale and age-related change, and the second mode captures facial slenderness coupled with head-shape variation. The PCA comparison reinforces that TRACE produces a more interpretable and clinically meaningful head shape space.

\section{Ablation Studies}
We perform ablation studies to evaluate the contribution of the coarse-to-fine
deformation cascade and the individual loss terms. All ablations use the DG-TRACE variant and the same training, validation, and test
protocol as the main experiments. We report the same surface-sampling,
topology-preservation, and SSM quality metrics used in the main quantitative
evaluation.

\begin{figure}
\centering
\includegraphics[width=\textwidth]{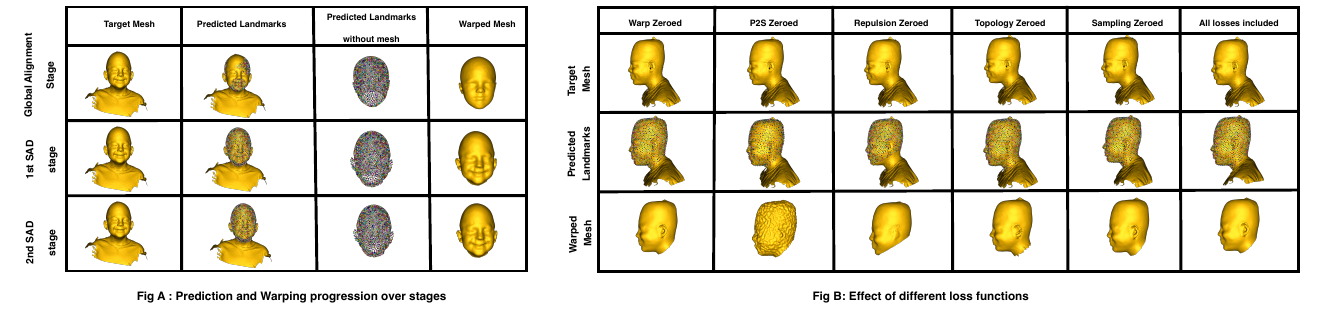}
\caption{\textbf{Qualitative ablation analysis:} \textbf{(A)} Prediction and warping progression across the deformation cascade. \textbf{(B)} Effect of removing individual loss terms to show the importance of each loss term.}
\label{fig:ablation}
\end{figure}

\begin{table}[t]
\centering
\caption{Ablation study of the deformation cascade for DG-TRACE variant. These variants differ only in the number of refinement stages applied after the global-alignment stage.}
\label{tab:stages}
\setlength{\tabcolsep}{4pt} 
\resizebox{\textwidth}{!}{
\begin{tabular}{l ccc ccc} 
\hline
& \multicolumn{2}{c}{\textbf{Surface Sampling}} & \multicolumn{1}{c}{\textbf{Topology Preservation}} & \multicolumn{3}{c}{\textbf{SSM Metrics }} \\
Method & P2F ($mm$) $\downarrow$ & S2S ($mm$) $\downarrow$ & Topo $\downarrow$ & Comp. $\downarrow$ & Gen. $\downarrow$ & Spec. $\downarrow$ \\
\hline
Global-Alignment Only  & 3.22  ± 0.88 & 3.32  ± 0.86 & \textbf{0.76 ± 0.34} & \textbf{4 }& \textbf{0.65} & \textbf{1.93} \\
Global-Alignment + 1 SAD  & 0.39 ± 0.06 & 0.74 ± 0.07 & 1.05 ± 0.49 & 11 & 2.60 & 3.97 \\
Global-Alignment + 2 SAD   & \textbf{0.29± 0.04}& \textbf{0.67 ± 0.05}& 1.10 ± 0.49 & 11 & 2.66 & 3.83 \\
\hline
\end{tabular}
}
\end{table}

\begin{table}[t]
\centering
\caption{Ablation study of training losses for DG-TRACE model. Each row removes one term from the full objective while keeping the same backbone, cascade, and training protocol. }
\label{tab:losses}
\setlength{\tabcolsep}{4pt} 
\resizebox{\textwidth}{!}{
\begin{tabular}{l ccc ccc} 
\hline
& \multicolumn{2}{c}{\textbf{Surface Sampling}} & \multicolumn{1}{c}{\textbf{Topology Preservation}} & \multicolumn{3}{c}{\textbf{SSM Metrics }} \\
Method & P2F ($mm$) $\downarrow$ & S2S ($mm$) $\downarrow$ & Topo $\downarrow$ & Comp. $\downarrow$ & Gen. $\downarrow$ & Spec. $\downarrow$ \\
\hline
P2S zeroed  & 2.24 ± 0.13 & 1.35 ± 0.08  & 1.57 ± 0.59 & 10 & 2.24 & 3.60 \\
Sampling zeroed  & 0.29 ± 0.043 & 0.67 ± 0.05 & 1.13 ± 0.50 & 12 & 2.92 & 4.39 \\
Topology zeroed   & 0.28 ± 0.03 & 0.68 ± 0.05 & 1.79 ± 0.47 & 13 & 3.10 & 3.93 \\
Warp zeroed   & 0.31 ± 0.04 & 0.69 ± 0.05 & 1.11 ± 0.51 & 11 & 2.65 & 3.74 \\
Repulsion zeroed   & \textbf{0.19 ± 0.02}& \textbf{0.57 ± 0.03}& \textbf{0.79 ± 0.32} & \textbf{9} & \textbf{2.09} & \textbf{3.52} \\
\hline
All losses included   & 0.29± 0.04 & 0.67 ± 0.05 & 1.10 ± 0.49 & 11 & 2.66 & 3.83 \\
\hline
\end{tabular}
}
\end{table}

\noindent{\textbf{Deformation cascade:}}
Table~\ref{tab:stages} evaluates the role of the global-alignment stage and the two SAD refinement stages. The global-alignment-only model preserves the
relative template structure, yielding the lowest topology error,  but fails to provide robust surface sampling, i.e., P2F and S2S remain high at 3.22 mm and
3.32 mm, respectively. As Fig.~\ref{fig:ablation}(A) shows, many landmarks remain under the target surface, and TPS output resembles a scaled template rather than a subject-specific reconstruction. Its strong compactness, generalization, and specificity therefore reflect a restricted template-like shape space rather than accurate sampling.
Adding the first SAD stage reduces P2F to 0.39 mm and S2S to 0.74 mm by moving landmarks toward the observed surface with a larger neighborhood. The second SAD stage further improves P2F to 0.29 mm and S2S to 0.67 mm using smaller local neighborhoods around the first-stage predictions, sharpening landmark locations and facial structure. As shown in Fig.~\ref{fig:ablation}(A), this coarse-to-fine refinement sharpens the final landmark locations and produces a warped mesh with clearer subject-specific
facial structure. The modest topology-error increase is expected because landmarks depart from the undeformed template to fit individual anatomy, while surface accuracy improves by more than an order of magnitude.

\noindent{\textbf{Loss terms:}}
Table~\ref{tab:losses} shows that the point-to-surface (P2S) loss is the dominant term for learning valid surface correspondences. Removing it increases P2F from the full-model range of roughly 0.29 mm to 2.24 mm and also degrades S2S, topology, and SSM quality, indicating that the remaining losses cannot by themselves anchor the landmarks to the head surface. This failure is also evident in Fig.~\ref{fig:ablation}(B), i.e., with the P2S term removed, the predicted landmarks are not consistently distributed, with most of them lying underneath the target surface, and the resulting warp lacks proper structure. Removing the topology loss produces the largest topology error of 1.79, and worsens SSM metrics, confirming that local template relationships are needed for anatomically ordered correspondences even when surface distances stay low. Removing sampling consistency has little effect on direct surface sampling but worsens generalization and specificity, while removing warp loss has a smaller effect. Removing the repulsion loss gives the best numerical scores across the reported aggregate metrics, but this exposes a limitation of the metrics rather than an improved correspondence model. Without repulsion, landmarks can concentrate on surface regions that minimize nearest-surface distances while undersampling peripheral anatomy such as the ears, lower jaw, and neck. This behavior improves the quantitative metrics that reward proximity and local edge consistency, but it reduces anatomical coverage, as shown in Fig.~\ref{fig:ablation}(B). If the objective were only to recover the central cranial vault or top-of-head structure, removing the repulsion term would be preferable under the reported aggregate metrics; however, for this study we prioritize anatomically broader head-surface coverage, including peripheral regions such as the ears, lower jaw, and neck, and therefore retain the repulsion term despite its slightly worse numerical scores. The full objective better balances surface anchoring, topology preservation, dense warping, and anatomical coverage. 

Overall, the ablations show that TRACE's main gains come from the SAD refinement cascade together with P2S anchoring and topology preservation, while sampling, repulsion, and warping terms primarily regulate stability, coverage, and dense mesh quality.

\section{Limitations and Future Work}
A limitation of the present study is the size of the held-out test set. Although the full dataset contains 201 clinical 3D photographs, the current evaluation
uses 25 test subjects due to available raw-scan constraints at this methods-development stage. Future work will therefore evaluate TRACE on larger, more diverse, multi-institutional cohorts. Clinical validation connecting SSM quality to expert-defined severity measures
is planned as the immediate follow-up study, pending a sufficiently large annotated multi-phenotype dataset. The prior work of Bruce et al.~\cite{bruce20233d} establishes that 3D photography contains sufficient information to reproduce CT-based severity scoring, and our work establishes the automated SSM foundation that makes this clinically scalable. Also, the normocephalic infant template used by TRACE provides a strong anatomical prior that may introduce bias for phenotypes that differ substantially from the template. In such cases, the deformation may favor preserving the template’s correspondence organization at the expense of accurately representing extreme or atypical morphology. The worst-performing example in Fig.~\ref{fig:qualitative}(B) provides a partial indication of this limitation. In addition, although TRACE captures the overall craniofacial geometry required for statistical shape modeling, fine facial details, particularly around the eyes and mouth, are not fully preserved. Future work will investigate methods to improve local correspondence and reconstruction fidelity in these anatomically complex regions.
Finally, while TRACE is demonstrated here on craniosynostosis 3D photographs, its design is anatomy-agnostic; evaluation on additional anatomies is planned as future work to validate this generality empirically.

\section{Conclusion}
We present TRACE, an unsupervised framework for constructing head-anatomy SSMs directly from raw clinical 3D photographs. By predicting sparse head-surface control points, refining them through a coarse-to-fine SAD cascade, and warping a clean template with TPS, TRACE constrains correspondence learning to anatomically meaningful head surfaces while reducing the influence of shoulders, hands, clothing, hair, and scanner artifacts. Experiments on clinical 3D scans show that this design consistently improves surface sampling, topology preservation, and SSM quality compared with prior deep learning-based SSM methods. The correspondence-prediction module is agnostic to the underlying point encoding backbone, and the resulting TRACE variants outperform prior SSM methods while producing cleaner, more interpretable population shape spaces. Overall, our findings establish TRACE as a scalable step toward objective craniosynostosis shape modeling from radiation-free 3D photography and support its use as a foundation for future longitudinal and multi-institutional severity-analysis workflows. Because the template is the only anatomy-specific element, the same framework applies to any anatomy for which a clean surface template exists, making it a general tool for SSM construction from imperfect clinical scans.

\begin{credits}

\subsubsection{{Ethics.}}
This study was conducted following approval by the Institutional Review Board (IRB; STUDY20110396). Data were obtained retrospectively from clinically acquired 3D photographs. Prior to use in this study, all images were deidentified and stripped of texture information to protect patient privacy and confidentiality.

\subsubsection{\ackname} 
The authors gratefully acknowledge the support of the National Institutes of Health under grant number NIDCR-R01DE032366. The content is solely the responsibility of the authors and does not necessarily represent the official views of the National Institutes of Health.
The authors also thank the members of the CranioRate Consortium, with special appreciation to the UPMC Children's Hospital of Pittsburgh for providing the data and valuable clinical expertise used in this study.

\subsubsection{\discintname}
The authors have no competing interests to declare that are relevant to the content of this article.

\end{credits}

%
%
%
\bibliographystyle{splncs04}
\bibliography{ref}

\end{document}